# INVESTIGATING SPHERICAL EPIPOLAR RECTIFICATION FOR MULTI-VIEW STEREO 3D RECONSTRUCTION

M. Elhashash[1,3], R. Qin[1,2,3,4,*]

[1] Geospatial Data Analytics Lab, The Ohio State University, Columbus, USA
[2] Department of Civil, Environmental and Geodetic Engineering, The Ohio State University, Columbus, USA
[3] Department of Electrical and Computer Engineering, The Ohio State University, Columbus, USA
[4] Translational Data Analytics Institute, The Ohio State University, Columbus, USA
Email: <elhashash.3><qin.324>@osu.edu

**Commission II – WG II/2**

**ABSTRACT:**
Multi-view stereo (MVS) reconstruction is essential for creating 3D models. The approach involves applying epipolar rectification followed by dense matching for disparity estimation. However, existing approaches face challenges in applying dense matching for images with different viewpoints primarily due to large differences in object scale. In this paper, we propose a spherical model for epipolar rectification to minimize distortions caused by differences in principal rays. We evaluate the proposed approach using two aerial-based datasets consisting of multi-camera head systems. We show through qualitative and quantitative evaluation that the proposed approach performs better than frame-based epipolar correction by enhancing the completeness of point clouds by up to 4.05% while improving the accuracy by up to 10.23% using LiDAR data as ground truth.

## 1. INTRODUCTION

Epipolar rectification is an essential step for multi-view stereo (MVS) 3D reconstruction. Assuming two images with known exterior orientation parameters, the epipolar rectification estimates homographic transformation for each of the two images, such that the transformed images have parallel epipolar lines in the row direction to facilitate dense stereo matching. In this case, these images are often corrected to fronto-parallel views (i.e., normal case of stereo photogrammetry), which is very suitable for traditional nadir aerial mapping. However, in oblique photogrammetry, the principal direction of images among these oblique images may form large angles, which may lead to large distortions when they are rectified to epipolar images. As a result, these distortions may impact the dense matching results.

Existing solutions seek different approaches to minimize these types of distortions (Liu et al., 2016; Oram, 2002). Nevertheless, the geometric distortion is challenging to quantify in many cases. Therefore, we propose to use a spherical model to perform the epipolar correction to minimize such distortions and investigate its potential to improve the MVS 3D reconstruction. Following the traditional pipeline for dense image matching, we propose to project the rectified epipolar images into a sphere and perform hierarchical semi-global matching (SGM) using the spherical images as inputs. The proposed approach is evaluated using two datasets collected with airborne multi-camera systems in two different areas and evaluated using LiDAR data as ground truth. To this end, our contributions are two-fold: firstly, we propose a spherical epipolar correction to reduce the distortions of the epipolar images. Secondly, the proposed approach enhances the completeness of the generated point clouds, while enhancing the accuracy by measuring the distance against LiDAR data as ground truth.

The rest of this paper is organized as follows: Section 2 reviews related works in MVS reconstruction; Section 3 introduces our proposed spherical-based rectification approach; Section 4 presents the experimental results, comparative studies, and analysis, and Section 5 concludes this work.

## 2. RELATED WORK

**Epipolar rectification.** The basic idea of performing epipolar rectification is to make the epipolar lines parallel to each other in the horizontal direction and thus reduce the search space while searching for correspondences. Several works have been introduced to perform epipolar rectification while aiming at minimizing epipolar distortion. (Roy et al., 1997) proposed to remap images on a cylinder instead of a plane. (Pollefeys et al., 1999) proposed a rectification algorithm that can handle forward motion using a polar parametrization. Other approaches (Liu et al., 2016; Oram, 2002) proposed a minimization framework to estimate the homography transformation while minimizing the perspective distortion effects. (Ohashi et al., 2017) proposed a rectification approach using two fisheye cameras, where the rectification parameters were estimated using feature points. In addition, (Wang et al., 2012) proposed a solution to work with an omnidirectional multi-camera system based on a spherical camera model.

**Dense image matching.** The main goal of dense image matching is to find correspondences between rectified image pairs. Traditional approaches search for correspondences by measuring the similarity between stereo pairs. The similarity is measured using different cost functions (Hirschmuller and Scharstein, 2009), such as Census (Zabih and Woodfill, 1994), which is one of the most robust cost matching functions. SGM (Hirschmuller, 2008) is regarded as one of the most reliable approaches for image matching. It creates a cost volume and finds matches that best minimize a cost function. A key challenge is its large memory requirement for computation (Qin, 2016). Thus, a hierarchical approach (Rothermel et al., 2012) has been proposed

---

[*] Corresponding author 2036 Neil Avenue, Columbus, Ohio, USA. qin.324@osu.edu

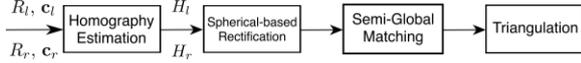
**Figure 1**. Overview of the proposed pipeline using spherical model correction.

to narrow the search range using an image pyramid scheme. PatchMatch-based (Barnes et al., 2009; Barnes et al., 2010) approaches are also widely used to solve the dense correspondence problem. In addition, learning-based approaches have been proposed leveraging the capabilities of convolutional neural networks (CNN) to extract reliable features. (Chang and Chen, 2018) proposed a pyramid stereo matching network and shown promising results.

## 3. METHODOLOGY

We introduce the main modules of the proposed approach. Given the exterior orientation parameters for two cameras, the goal is to estimate a homography transformation to ensure the epipolar lines are parallel. After estimating the homography transformations, the images are rectified and we apply the spherical model to project the rectified images into a spherical image. The spherical images are used for disparity estimation through an SGM-based algorithm. The pipeline of the proposed method is shown in **Figure 1**.

### 3.1 Homography Estimation

As mentioned in Sections 1 and 2, to reduce the search space for image matching, we define a transformation that maps the epipoles to infinity along the horizontal axis. Thus, epipolar lines will be parallel to each other and correspondences between two images will be in the same rows (i.e., there is only a translation along the X-axis). We begin by defining the projection model of an object point **X** onto the left and right frame-based images, as follows:

$$\begin{aligned}\mathbf{x}_l &= K[R_l\ -R_l\mathbf{c}_l]\mathbf{X} \\ \mathbf{x}_r &= K[R_r\ -R_r\mathbf{c}_r]\mathbf{X}\end{aligned} \quad (1)$$

where K is the intrinsic parameters, while $R_l$, $\mathbf{c}_l$ and $R_r$, $\mathbf{c}_r$ denote the extrinsic parameters of the left and right cameras, respectively. Note for simplicity, we assume both cameras have the same intrinsic parameters, whereas arbitrary new intrinsic parameters can be estimated if they are not the same (e.g., an average of both). After rectification, both cameras should have the same rotation matrix $R = [\mathbf{r}_1\ \mathbf{r}_2\ \mathbf{r}_3]^T$. We define the rows of R as follows

1. The X-axis is parallel to the baseline: $\mathbf{r}_1 = (\mathbf{c}_r - \mathbf{c}_l)/\|\mathbf{c}_r - \mathbf{c}_l\|$.
2. The Y-axis is defined as: $\mathbf{r}_2 = \mathbf{r}_1 \times \mathbf{k}$, where $\mathbf{k} = \mathbf{r}_{3l} + \mathbf{r}_{3r}$ (i.e., the sum of the principal rays of both cameras).
3. The Z-axis is defined as: $\mathbf{r}_3 = \mathbf{r}_1 \times \mathbf{r}_2$.

After formulating the new rotation matrix R, the homography transformation between the original image and the rectified one can be estimated using $H_l$ and $H_r$, such that:

$$\begin{aligned}H_l &= KRR_l^{-1}K^{-1} \\ H_r &= KRR_r^{-1}K^{-1}\end{aligned} \quad (2)$$

### 3.2 Spherical Model Epipolar Correction

After estimating the homography transformations $H_l$ and $H_r$ as explained in Section 3.1, we apply them to the input stereo image pair to obtain epipolar rectified images. Then, we project the epipolar images into a sphere. The rationale is that the spherical

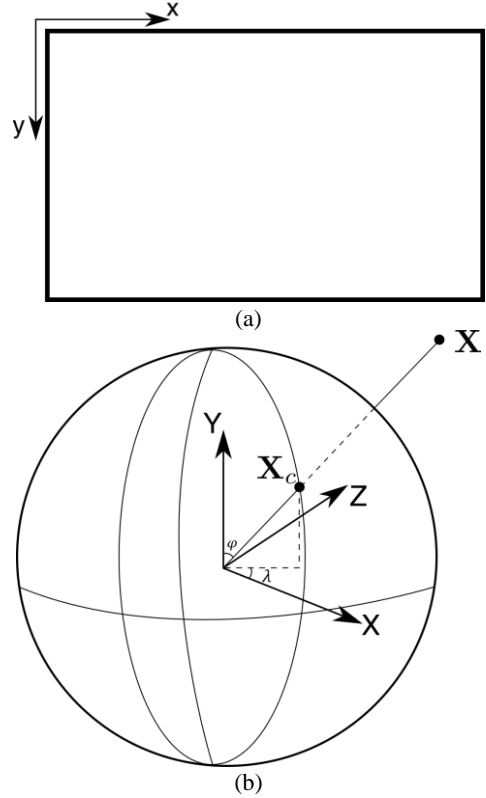
**Figure 2**. Coordinate systems of frame-based image in (a), and spherical in (b).

image reduces the distortions caused by the large changes in the principal rays between a pair of left and right images. Thus, we define the mapping used between a frame-based image and its corresponding spherical image. As shown in **Figure 2**, the sphere is parameterized by two angles $(\lambda, \varphi)$ which denotes the longitude and latitude. The projection of a pixel in the frame-based image $(x, y)$ to the point $\mathbf{X}_c$ in a sphere can be defined as:

$$\begin{aligned}X_c &= \frac{\left(x - \frac{w}{2}\right)}{f} \\ Y_c &= \frac{\left(y - \frac{h}{2}\right)}{f} \\ Z_c &= 1\end{aligned} \quad (3)$$

where $(X_c, Y_c, Z_c)$ denotes the normalized coordinate of $(x, y)$ in the camera coordinate system, $w$ and $h$ are the width and height of the frame-based image, respectively. Thus, we can define the angles $(\lambda, \varphi)$ as follows:

$$\begin{aligned}\lambda &= \tan^{-1}\left(\frac{X_c}{Z_c}\right) \\ \varphi &= \tan^{-1}\left(\frac{-Y_c}{r}\right) \\ r &= \sqrt{X_c^2 + Z_c^2}\end{aligned} \quad (4)$$

Since $Y_c$ (which is the same in two frame-based epipolar images) is divided by $r$ which depends on $X_c$, if we apply Eq. (4), rows in two epipolar images (or $\varphi$ in the spherical coordinates) will not be the same because $X_c$ is not the same for two epipolar images. Thus, we swap $X_c$ and $Y_c$ before applying Eq. (4) so that $\lambda$ will

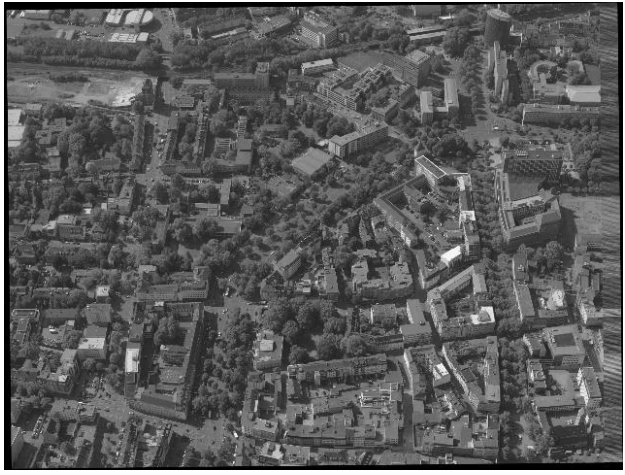 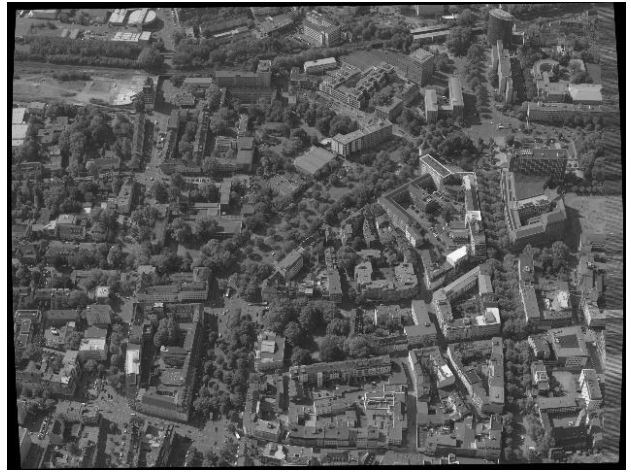

(a) Frame-based epipolar image  (b) Spherical epipolar image

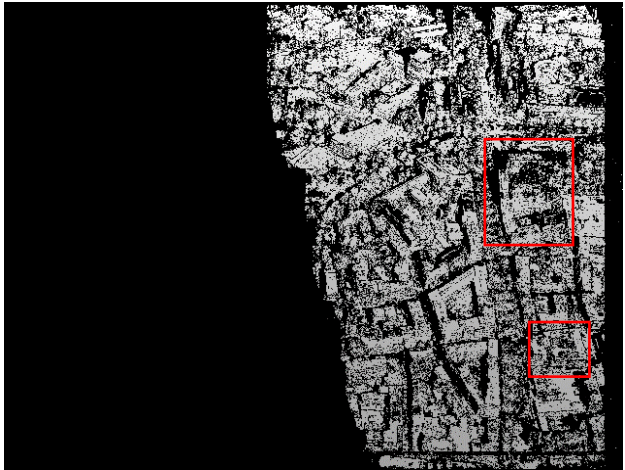 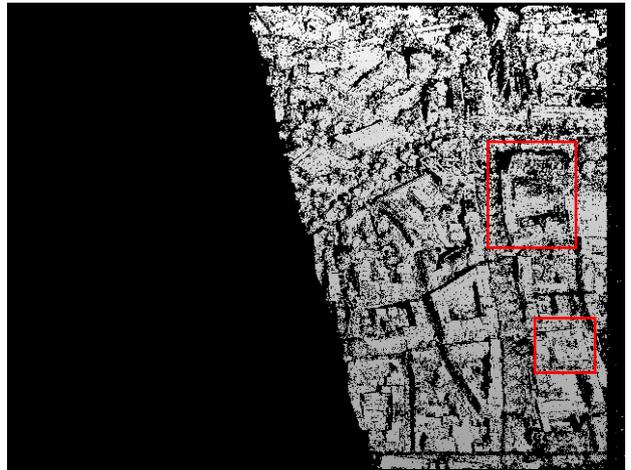

(c) Depth map generated using frame-based model  (d) Depth map generated using spherical-based model

**Figure 3**. Comparison of depth maps computed using the proposed spherical model compared to frame-based rectification. Top row shows the epipolar images. The outlined regions show examples of a more complete depth map generated by the proposed spherical model compared to frame-based rectification.

be the same for two epipolar images. Then, after applying Eq. (5), we rotate the warped spherical image 90° clockwise. The coordinate of a pixel in the spherical image can be obtained by:

$$u = \frac{s\lambda}{2\pi}$$
$$v = \frac{s\varphi}{2\pi}, \quad (5)$$

where $(u, v)$ is the pixel position, $s$ is a scale parameter, and $\frac{s}{2\pi}$ defines the number of pixels/angle for the spherical image.
Given $(\lambda, \varphi)$, the mapping from the spherical image to the camera coordinate system is defined as follows:

$$X_c = \sin\lambda \cos\varphi$$
$$Y_c = -\sin\varphi \quad (6)$$
$$Z_c = \cos\lambda \cos\varphi,$$

which can be projected to the frame-based image as follows:

$$x = f\frac{X_c}{Z_c} + \frac{w}{2}$$
$$y = f\frac{Y_c}{Z_c} + \frac{h}{2}. \quad (7)$$

### 3.3 Hierarchical Semi-Global Matching

We use SGM (Hirschmuller, 2008) as the matching strategy for stereo pairs. The method applies multi-path dynamic programming to optimize a cost function that consists of pixel-wise cost and smoothness constraints to penalize the change of neighboring disparities. This cost function is defined as follows:

$$E(D) = \sum_p C(p, D_p) + \sum_{q \in N_p} P_1 T[|D_p - D_q| = 1]$$
$$+ \sum_{q \in N_p} P_2 T[|D_p - D_q| > 1] \quad (8)$$

The first term represents the sum of matching cost for the disparities of $D$, while the second and the third terms control the smoothness of the disparity map. $N_p$ is the set of neighboring pixels of $p$, and $T[.]$ is a logical operator that enables or disables the error term according to its condition. $P_1$ denotes the penalty value for a small disparity change of 1 pixel, and $P_2$ penalize disparity changes of more than 1 pixel, which often occurs at the surface discontinuities. We use Census cost (Zabih and Woodfill, 1994) since it is robust against illumination changes. Typically, the traditional SGM creates a raster to store the aggregated cost for each disparity value, which requires a large volume of memory for computation. Therefore, we applied a hierarchical solution to run the SGM algorithm through a pyramid of the image using the algorithm described in (Qin, 2016; Rothermel et al., 2012). **Figure 3** shows a sample of the computed depth maps using the proposed approach compared to frame-based rectification. We can visually observe the high number of points obtained by the proposed approach.

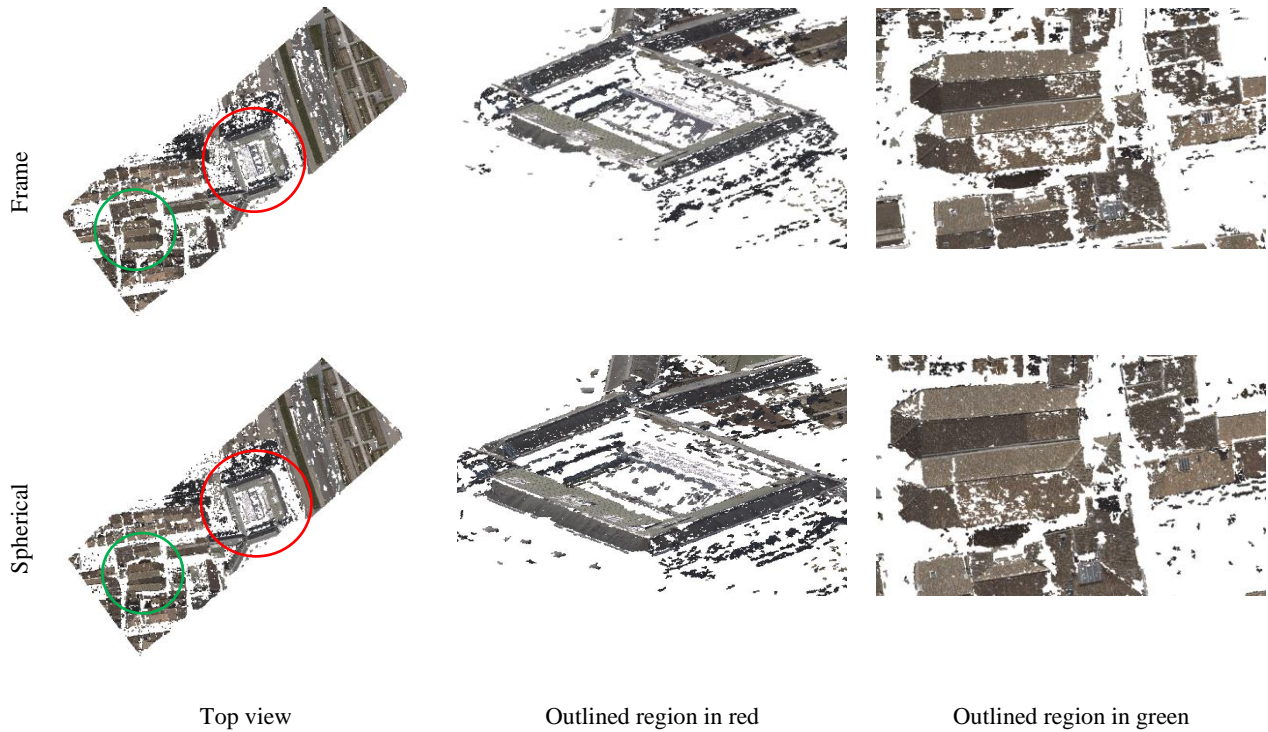

**Figure 4**. Generated point clouds using the frame-based rectification (first row) compared to the proposed approach using the proposed spherical model. The circled regions give examples of the more complete point cloud produced using the proposed approach compared to frame-based.

### 3.4 Triangulation

The last step is to triangulate the correspondences obtained using the disparity estimation by SGM. Given the disparity value for a pair of correspondences in the left and right images that are located in the same row, the depth $Z$ can be estimated as $Z = \frac{bf}{d}$, where $f$ is the focal length, $b$ is the baseline (distance between two cameras), and $d$ is the disparity value.

## 4. EXPERIMENTAL RESULTS

We evaluated the proposed approach on two aerial datasets. The produced point clouds are evaluated qualitatively and quantitatively in terms of completeness (i.e., number of points) and accuracy using LiDAR data as a reference. For all the reported experiments, the matching pairs are selected according to the number of SIFT (Lowe, 2004) points. We perform dense matching for each image against ten images and evaluate the results for both frame-based and the proposed spherical model epipolar correction. Note that our goal is not to evaluate the pair selection criteria but the ability of the proposed approach to produce complete and accurate point clouds. Section 4.1 describes the datasets used in the experiments. The qualitative and quantitative evaluations of the generated point clouds are provided in Section 4.2 and Section 4.3, respectively.

|  | Dortmund | Bordeaux |
|---|---|---|
| Number of images | 16 nadir (N)<br>43 oblique (O) | 33 nadir (N)<br>35 oblique (O) |
| Camera | Pentacam IGI | Leica CityMapper |
| Image resolution | 6132 x 8176 px (N)<br>8176 x 6132 px (O) | 10336 x 7788 px (N)<br>10336 x 7788 px (O) |
| Focal length | 50 mm (N)<br>80 mm (O) | 83 mm (N)<br>156 mm (O) |
| Pixel size | 6 mm (N)<br>6 mm (O) | 5.2 mm (N)<br>5.2 mm (O) |
| Platform | airborne | airborne |
| Overlap (N) | 75/80 | 80/60 |
| Avg GSD | 10 cm (N)<br>8-12 cm (O) | 5 cm (N) |

**Table 1**. Properties of the datasets used in the evaluation.

### 4.1 Dataset

We utilize two datasets to evaluate the proposed approach captured in the city of Dortmund, Germany, and Bordeaux, France. Both datasets were acquired by a multi-camera system in an airborne platform. We summarize the properties of the datasets used in the evaluation in **Table 1**. The Dortmund data is from the ISPRS/EuroSDR Dortmund benchmark (Nex et al., 2015). The data was captured by IGI PentaCam camera system. We use 43 oblique images, and 16 nadir images in our experiments. The GSD is 10 cm for nadir images and ranges from 8 to 12 cm in the oblique images. The overlap is 75%/80% (along/across-track directions) for the nadir images while it is 80%/80% for oblique images. The Bordeaux data was collected using Leica CityMapper hybrid sensor. The nadir images have an overlap of 80% and 60% and were collected using a Leica RCD30 CH82 multispectral camera and the oblique images were collected using four Leica RCD30 CH81 cameras. The nadir images have a mean nadir GSD of 5 cm. We use 33 nadir images, and 35 nadir images in our experiments.

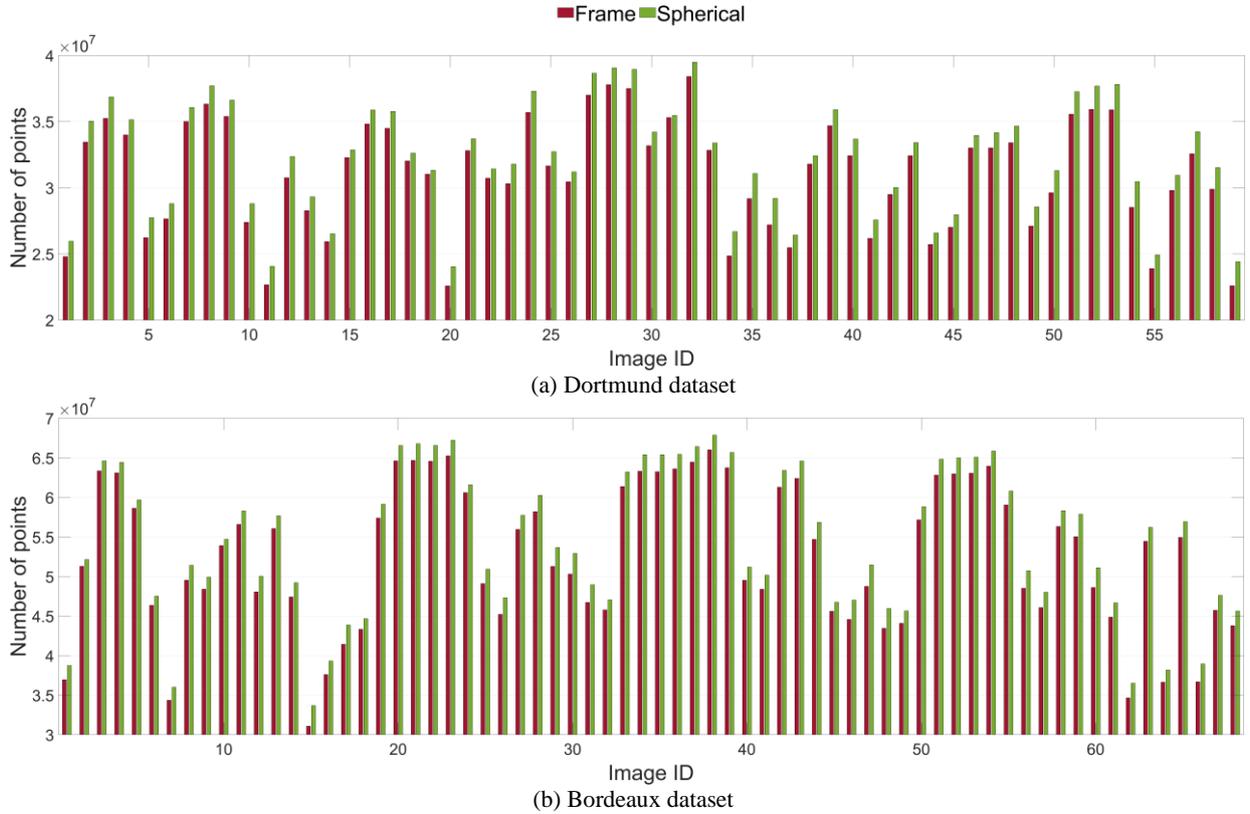

**Figure 5**. Evaluation of the completeness of generated point clouds for the Dortmund and Bordeaux datasets using the proposed spherical matching compared to frame-based.

### 4.2 Qualitative Evaluation

To evaluate the generated point cloud, we provide a comparison between the proposed spherical model correction and the frame-based rectification. A visual comparison is shown in **Figure 4** for a sample of the results. We can observe that the proposed approach enhances the completeness of the generated point clouds in different cases compared to frame-based rectification.

### 4.3 Quantitative Evaluation

As previously mentioned, we evaluated our approach using two metrics:
- The number of generated points to evaluate the completeness (**Figure 5**).
- The cloud-to-cloud distance to LiDAR data to measure the accuracy (**Table 2**). The distance computation is realized using CloudCompare open-source software (Girardeau-Montaut, 2015). The distance is measured after registering the generated point cloud to the LiDAR data.

|         | Method    | Dortmund | Bordeaux |
|---------|-----------|----------|----------|
| Mean (m)| Frame     | 0.879    | 0.309    |
|         | Spherical | 0.789    | 0.285    |
| Std (m) | Frame     | 0.769    | 0.233    |
|         | Spherical | 0.748    | 0.234    |

**Table 2**. Mean absolute distance and standard deviation of the cloud-to-cloud evaluation for Dortmund and Bordeaux datasets.

**Completeness.** We evaluate the completeness of the generated point clouds for the two datasets used for experiments. The results are shown in **Figure 5** for both datasets. The proposed approach produced on average 4.05% and 3.70% more points than the frame-based for Dortmund and Bordeaux data, respectively. Thus, the proposed approach outperforms frame-based rectification in both datasets in terms of completeness.

**Accuracy.** We use airborne LiDAR data as ground truth to evaluate the accuracy of the generated point clouds. We measure the mean absolute distance and the standard deviation of the distance between the generated point clouds using the frame-based rectification compared to the proposed approach. The results are shown in **Table 2** for both datasets. We can observe that the proposed epipolar correction using the spherical model achieves an improvement of 10.23% for the Dortmund dataset (reduced from 0.879 to 0.789 meters) and an improvement of 7.6% for the Bordeaux dataset (reduced from 0.309 to 0.285 meters). Thus, we can conclude that the proposed approach effectively produces not only complete but also accurate point clouds.

## 5. CONCLUSION

In this paper, we have presented a spherical model approach for epipolar correction to improve MVS 3D reconstruction. The proposed approach facilitates the dense image matching process by minimizing the geometric distortions caused by variations between viewpoints of stereo pairs. We validated the proposed method using two datasets consisting of nadir and obliques images. The proposed approach has achieved better results compared to frame-based rectification in terms of completeness and accuracy on both datasets. Our approach can be easily integrated into existing pipelines and help produce more points and better accuracy.

## ACKNOWLEDGMENTS

The authors would like to acknowledge the provision of the Dortmund dataset by ISPRS and EuroSDR. The authors would also like to acknowledge the provision of the Bordeaux dataset by Fondazione Bruno Kessler. This work was supported in part by the Office of Naval Research (Award No. N000141712928 and N000142012141).